\documentclass[journal]{IEEEtran}
\usepackage{authblk}
\usepackage{cite}
\usepackage{color}
\usepackage{float}
\usepackage{caption}
\usepackage{graphicx}
\usepackage{tabularx}
\usepackage{array}
\usepackage{adjustbox}
\newcolumntype{P}[1]{>{\centering\arraybackslash}p{#1}}
\usepackage{hhline}
\usepackage{makecell}
\usepackage{adjustbox}
\usepackage[utf8]{inputenc}
\usepackage[margin=1in]{geometry}
\usepackage{multicol}
\usepackage{multirow}
\usepackage{array}
    \newcolumntype{P}[1]{>{\centering\arraybackslash}p{#1}}
    \newcolumntype{M}[1]{>{\centering\arraybackslash}m{#1}}

\usepackage{caption}

\usepackage[T1]{fontenc}
\usepackage[table]{xcolor} 
\usepackage{dblfloatfix}

\usepackage{longtable}

\usepackage{tikz}
\usetikzlibrary{trees}

\ifCLASSINFOpdf
\else
\fi
\begin{document}
%
\title{Blockchain and Biometrics: Survey, \\GDPR Analysis, and Future Directions}
%
%
\author[1]{Mahdi~Ghafourian}
\author[2]{Bilgesu~Sumer}
\author[1]{Ruben~Vera-Rodriguez}
\author[1]{Julian~Fierrez}
\author[1]{Ruben~Tolosana}
\author[1]{Aythami~Morales}
\author[2]{Els~Kindt}
\affil[1]{Biometrics and Data Pattern Analytics Lab, Universidad Autonoma de Madrid, Spain}
\affil[2]{Biometric Law Lab, Centre for IT \& IP Law (CITIP), KU Leuven, Belgium}
\markboth{}%
{Shell \MakeLowercase{\textit{et al.}}: Bare Demo of IEEEtran.cls for IEEE Journals}
%



\maketitle

\begin{abstract}
Biometric recognition as an efficient and hard-to-forge way of identification and verification has become an indispensable part of the current digital world. The fast evolution of this technology has been a strong incentive for integration into many applications. Meanwhile, blockchain, the decentralized ledger technology, has been widely received by both research and industry in the past few years, and it is being increasingly deployed today in many different applications, such as money transfer, IoT, healthcare, or logistics. Recently, researchers have started to speculate on the pros and cons and what the best applications would be when these two technologies cross paths. This paper provides a survey of the research literature on the combination of blockchain and biometrics and includes a first legal analysis of this integration based on GDPR to shed light on challenges and potentials. Although the integration of blockchain technology into the biometric sector is still in its infancy, with a growing body of literature discussing specific applications and advanced technological setups, this paper aims to provide a holistic understanding of blockchain applicability in biometrics. Based on published studies, this article discusses, among others, practical examples combining blockchain and biometrics for novel applications in PKI systems, distributed trusted services, and identity management. Challenges and limitations when combining blockchain and biometrics that motivate future work will also be discussed; e.g., blockchain networks at their current stage may not be efficient or economical for some real-time biometric applications. Finally, we also discuss key legal aspects of the EU General Data Protection Regulation (GDPR) related to this combination of technologies (blockchain and biometrics); for example, accountability, immutability, anonymity, and data protection elements.
\end{abstract}


%
\IEEEpeerreviewmaketitle

\section{Introduction}
%
%
%
%
\IEEEPARstart{U}{nlike} conventional authentication methods based on knowledge (e.g. password) or possession (e.g. smart card), biometric recognition relies on the concept of inherence, i.e. who someone is, which makes it more robust against fraudulent activities like forging, spoofing, etc., compared to the aforementioned methods \cite{Ref3}. The advantages of biometric recognition include a high level of security \cite{2021_Fierrez}, a better user experience \cite{2020_CDS_HCIsmart_Acien}, and a fast recognition process that have allowed the use of biometric systems to improve many different applications, such as authentication systems \cite{2020_ICIP_QIDmulti_Perera}, border control \cite{2023_IET-Biom_Schengen_Busch}, etc.\cite{Ref4}. One of the technologies with great potential to be used in conjunction with biometrics is blockchain.

The advent of Distributed Ledger Technology (DLT), and in particular blockchain, has become one of the most prominent phenomena of our time. As a result, it has come to the forefront of research in many disciplines (e.g., cryptocurrency, smart contracts, identity management) aimed at including distributed data storage. Following the success of Bitcoin and other cryptocurrencies, the underlying blockchain technology is now widely used to distribute data through secure-by-design databases without the need for a central authority. Not only have the unprecedented security properties of blockchain revolutionized a wide range of financial services and digital payments, but its unique characteristics such as decentralization, immutability, auditability, fault tolerance, and availability have played a pivotal role in receiving further public attention so far.


On the one hand, it is not surprising that, due to its availability and decentralization properties, the blockchain solves the security problems arising from storing biometric features (e.g., storing biometric features of the same individual in different places belonging to independent applications). In particular, in terms of biometric verification, the so-called on-chain storage of biometric templates helps to prevent Denial of Service (DoS) attacks caused by a single point of failure. However, a major breakthrough for the combination of blockchain and biometrics would be the introduction of the Public Key Infrastructure (PKI) mechanism\cite{maurer1996modelling}. Owing to their inherent connection to an individual's identity, biometrics lack PKI structure. To address this limitation, the advantages of using blockchain and more specifically a novel introduced concept called distributed identifiers (DIDs) are being investigated. Although current studies are almost focused only on the role of the blockchain in biometric verification, identity management, and e-voting, we discuss the unprecedented merits of combining blockchain and biometrics for other potential applications such as using blockchain as computing power for biometrics or introducing a foundation for PKI mechanism. In addition, we discuss the role of blockchain in enabling multimodal biometrics, providing self-Sovereign Identity (SSI), and serving as a computational medium in Proof-of-Work (PoW)-based consensus for biometric verification.

However, the shortcomings of combining blockchain and biometrics need to be analyzed specifically with regard to issues such as the type and location of information that can be stored on blockchain. One of the main objectives of this work is to analyze the performance of the system, for example, latency, computation cost, interoperability, and user convenience.

Another objective of this study is to provide a first legal and data protection analysis according to the General Data Protection Regulations (GDPR). Since May 25, 2018, the GDPR has been applicable in all EU member states to
harmonize data privacy laws across Europe. According
to the GDPR, any information related to an identified and
identifiable person is defined as personal data and shall
be subject to the provisions of the GDPR.  Biometric data stored on the blockchain in many applications are considered personal data, which is subject to GDPR. This paper
discusses the implications based on GDPR of storing
biometric data on public blockchains.


  In particular, the main contributions of this paper are as follows.
\begin{itemize}
  \item An overview of the meeting points of blockchain and biometrics with respect to both technical and legal potentials and limitations. 
  \item A summary and categorization of the literature from the perspective of key factors in applying blockchain to biometrics.
  \item An in-depth analysis of typical application scenarios for combining blockchain and biometrics while taking into account well-known aspects of blockchain such as scalability, smart contracts, consensus algorithms, security and privacy \cite{privacy2025}, and legal issues.
  \item A description of the unprecedented use cases for combining blockchain and biometrics and the corresponding merits.
\end{itemize}


In this regard, Section \ref{sec:def} gives a concise introduction to blockchain technology, biometrics, and their combination. In Section \ref{sec:main}, we analyze the impact of blockchain features when combined with biometrics from a technical point of view. This section discusses the prospects and limits that lie ahead of this combination. In this section, we also provide a categorization of the technical literature on combining blockchain and biometrics to improve biometric systems. Section \ref{sec:Legal_review} contains the legal analysis based on GDPR. Section \ref{sec:rec} offers some recommendations and potential research directions for future studies in this field. Finally, Section \ref{sec:con} concludes this survey.

\section{Blockchain and Biometrics: Introduction}\label{sec:def}
\subsection{Blockchain Technology}
\begin{figure*}[htp]
 \centering 
 \includegraphics[trim={0cm 5.5cm 0 7cm},clip,width=170mm,scale=0.5]{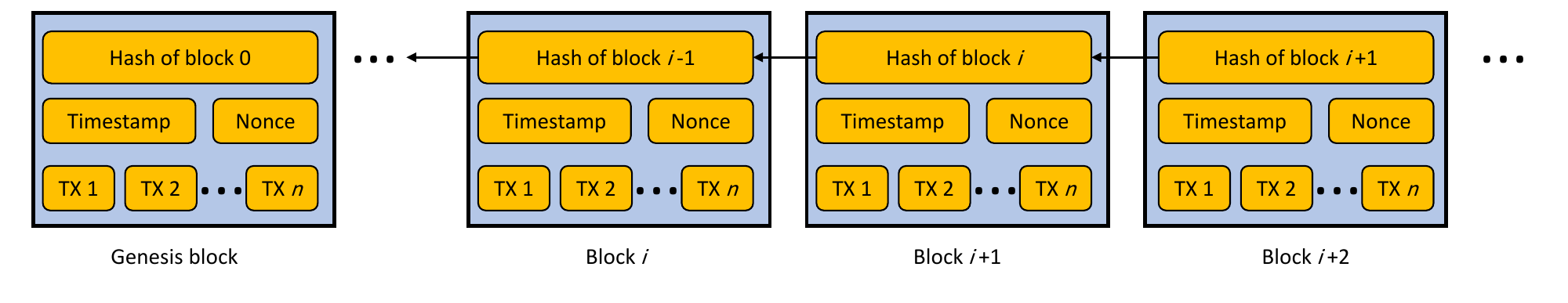}
 \caption{An example of a blockchain structure consisting of a continuous sequence of blocks taken from \cite{zheng2017overview}.}
 \label{fig:structure}
\end{figure*}

Blockchain is a distributed ledger shared among a group of entities that are supposed to reach a consensus to register or modify transactions \cite{yli2016current, pilkington2016blockchain, zheng2017overview}. These transactions are secure, immutable, and easy to track and measure. In particular, blockchain allows entities to reach an agreement on a certain subject and records it as a transaction without the need for a central authority. The subject of transactions varies depending on the application in which the blockchain is used. The structure of a common blockchain network is shown in Fig.~\ref{fig:structure}.

In particular, the header contains the following: \textit{the owner identifier} and the \textit{identifier of the previous block}, a \textit{timestamp} denoting the generation time of the current block, a \textit{nonce} (a number used in PoW to find a valid block hash, or in transactions to ensure uniqueness and ordering), and a \textit{merkle tree} root hash which denotes the hash value of all transactions stored in the block. The body consists of transactions and a transaction counter. The maximum number of transactions that fit within the body depends on the block size and the size of the transactions within. A digital signature based on asymmetric cryptography is used to validate transactions \cite{zheng2017overview}.

The most prominent characteristics of blockchain that have made it highly popular for integration with other technologies, including biometrics, are \textit{decentralization}, \textit{immutability}, \textit{auditability}, \textit{fault tolerance}, and \textit{availability} \cite{christidis2016blockchains, 2020_COMPSAC_BlockIOT_Oscar}.

\begin{figure*}[bp]
 \centering 
 \includegraphics[trim={3cm 3cm 2cm 4cm},clip,width=170mm,scale=0.5]{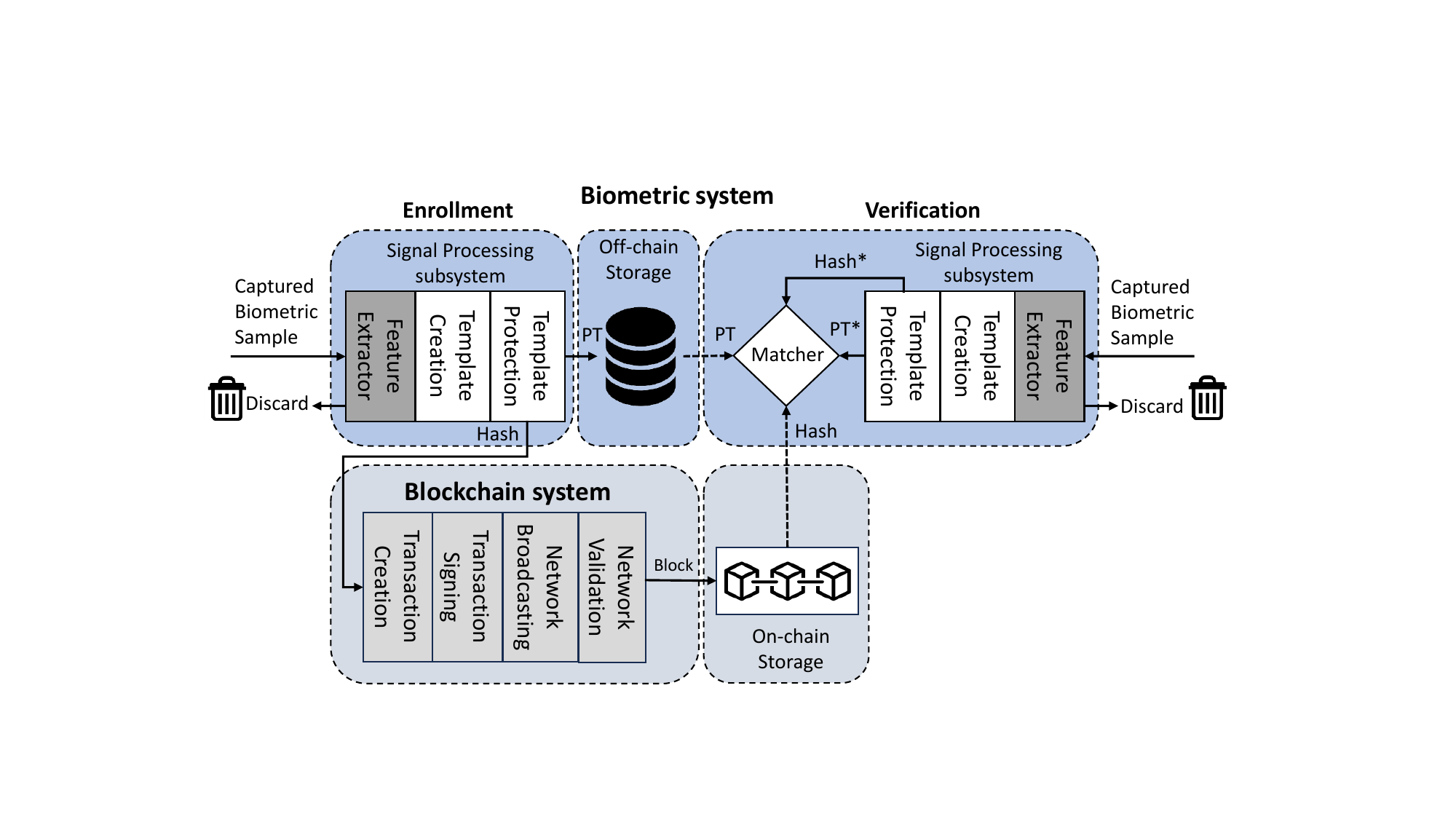}
 \caption{The lifecycle of a biometric transaction on the blockchain (PT = Protected Template).}
 \label{fig:lifecycle}
\end{figure*}

\subsection{Biometric Technology}
Biometrics in modern computer science refers to the automated utilization of biological traits for personal identification. These traits enable humans to distinguish individuals based on their physical (e.g. hand geometry, fingerprints, face, iris, and so on) and behavioral (e.g. signature, keystroke, and etc.) characteristics. When properly implemented, they also allow computer systems to recognize patterns for security purposes \cite{bharathi2019review}. The key phases of a biometric system are enrollment and authentication. These phases involve several essential processes: (i) sensor modules for modality acquisition, (ii) feature extraction in the feature module, (iii) comparison of extracted features with the database in the matching module, and (iv) template verification in the decision module. At the sensor level, biometric modalities are captured as images, videos, audio, or signals. The extracted biometric characteristics are represented as a set of points or vectors. The acquired modalities are stored as templates for future authentication, verification, and identification. Biometric features extracted from biometric samples are stored as a template, which is therefore called a biometric template \cite{ortiz2018survey}.

\subsection{Blockchain for Biometrics}
Blockchain and biometric systems can be integrated in various ways. Here, we discuss the process involved in a typical blockchain transaction intersected with the typical biometric workflow. 

A transaction on the blockchain represents an interaction between the parties \cite{yaga2019blockchain}. In the context of biometrics, recording the result of an authentication is considered a transaction. Common biometric operations, such as template creation, modification, deletion, and retrieval, are also considered transactions that are recorded on the blockchain\cite{Ref68}. The lifecycle of a biometric transaction on the blockchain is shown in Fig. \ref{fig:lifecycle}.

This workflow begins when a user provides their biometric input, such as a fingerprint or facial scan, using a secure device. This raw biometric data is then processed to create a unique digital representation, known as a biometric template. To ensure privacy, the template is secured using the robust protection method. The protected template (PT) can be stored off-chain or on-chain. Given the sensitive nature and size of biometric data, it is typically stored in a secure off-chain database, while a cryptographic hash of the encrypted template is generated and stored on the blockchain. This hash serves as a reference to verify the integrity of the biometric data without exposing the actual template. A blockchain transaction is then created, including this cryptographic hash, and signed using the user's private key to ensure authenticity. The signed transaction is transmitted to the blockchain network, where the nodes validate it by checking the digital signature and ensuring that it adheres to the protocol rules. Once validated, the transaction is included in a new block and added to the blockchain through the network consensus mechanism. For future verification, the user provides their biometric input again, a new template is generated and encrypted, and the system compares the hash of this new template to the one stored on the blockchain to confirm identity.

\begin{figure}[tbp]
 \centering 
 \includegraphics[trim={6.8cm 0.5cm 1cm 1cm},clip,width=100mm,scale=0.5]{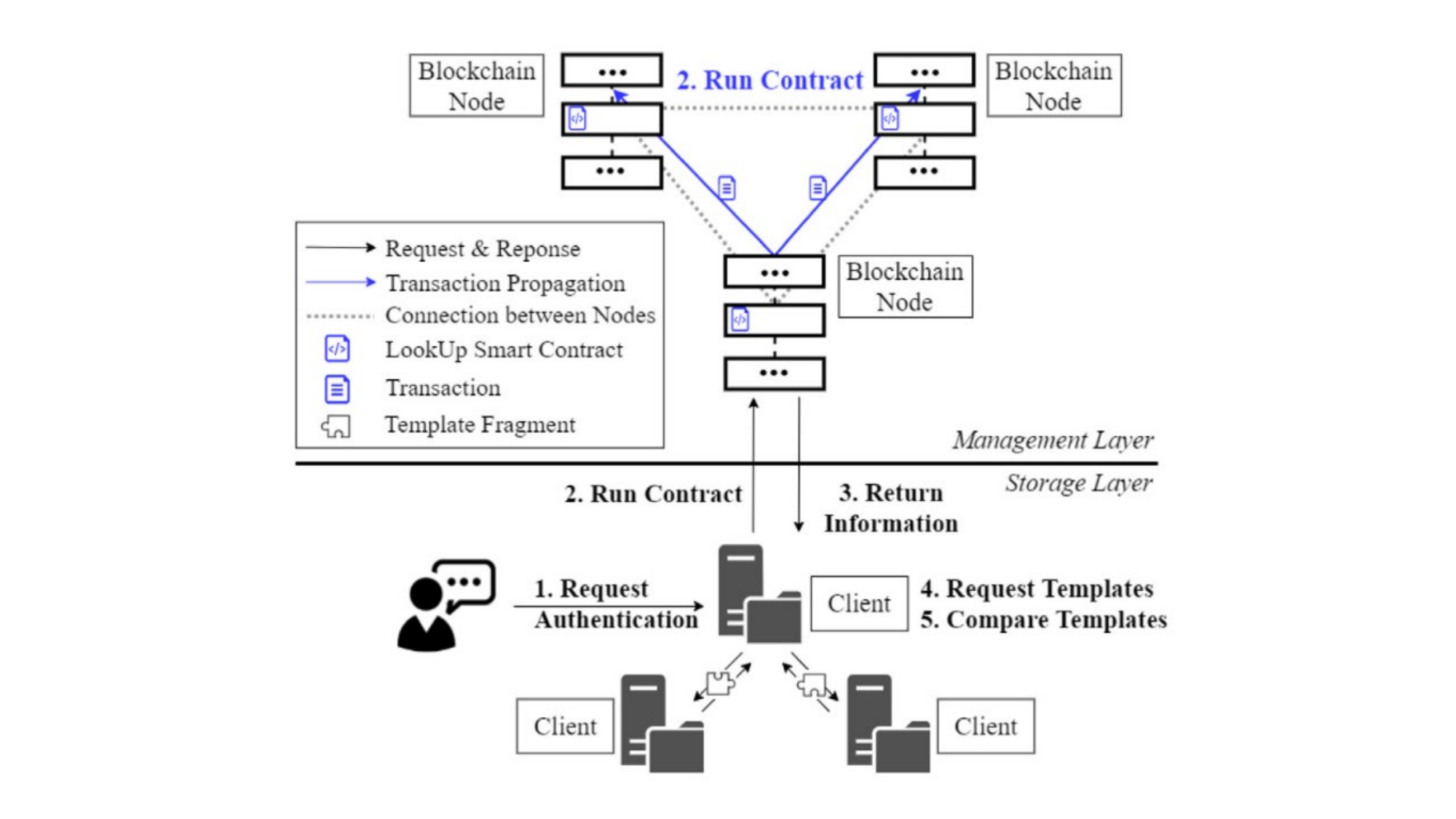}
 \caption{An overview of a biometric authentication system using blockchain. (Image taken from\cite{Ref28}.)}
 \label{fig:BDAS}
\end{figure}

Integrating biometrics with blockchain technology encompasses various applications and can impact multiple modules within biometric systems. For example, Delgado-Mohatar et al.\cite{delgado2020blockchain} explored implementing each component of a biometric template protection system on a blockchain. Biometric authentication is the most common application of biometrics. In this context, Lee and Jeong\cite{Ref28} proposed an end-to-end protocol consisting of enrollment and authentication called \textit{BDAS}, based on blockchain, as shown in Fig.~\ref{fig:BDAS}.

The \textit{BDAS} enrollment process involves (1) capturing user biometric data to create a template, (2) splitting this template into three fragments based on its timestamp, (3) identifying connected nodes, (4) selecting clients to store these fragments, each fragment stored in \(\left\lfloor \frac{n}{3} \right\rfloor\) copies among \(\left\lfloor 3 \times \left( \frac{n}{3} \right) \right\rfloor\) clients, and (5) storing the assigned template fragments in the selected clients.

The authentication process comprises five steps: (1) a client captures biometric data to request authentication, (2) runs a LookUp smart contract to obtain the locations of the required template fragments, which (3) are returned; (4) the client retrieves these fragments through separate communications, (5) merges them into a complete template, and compares it with the stored data.

\subsection{Related Works}
Several studies are related to the current survey. In 2022, Ahmed et al. \cite{ahmed2022blockchain} provided an in-depth analysis of how blockchain technology is revolutionizing identity management systems (IdMs), particularly through the lens of SSI. Their study explored the SSI ecosystem, emphasizing the role of decentralized identifiers and verifiable credentials in empowering users with control over their personal data.

In the same year, Rathee and Singh \cite{rathee2022systematic} analyzed how blockchain technology can enhance IdMs by addressing security and privacy concerns. Their study showed that blockchain-based identity systems outperform traditional centralized systems in terms of transparency, security, and ease of use for both individuals and organizations. Similarly, Ngo et al. \cite{ngo2023systematic} provided a systematic overview of blockchain-based IdMs, classifying them according to their contributions, application domains, identity types, and research methodologies. Their study focused on papers from four databases (IEEE Xplore, ScienceDirect, ACM, and Springer Link) published between 2009 and 2022, and proposed general solutions for the application of blockchain to identity management.

The study that aligns the most closely with the objective of our survey is the work published by Sharma and Dwivedi \cite{sharma2024survey}. They examined various applications of blockchain in biometric systems, including biometric template storage, identity management, and authentication processes. However, their analysis can be criticized for its superficial treatment of the subject matter, lacking depth in its exploration of the technical and legal complexities involved.

Although Rathee and Singh \cite{rathee2022systematic} mentioned Civic \cite{civic}, which claims to be GDPR compliant, all these related works lack any assessment of legal or GDPR-based regulations regarding the combination of blockchain and biometrics.

\section{Blockchain and Biometrics: Advantages and Limitations} \label{sec:main}
\begin{figure*}[tp]
 \centering 
 \includegraphics[trim={0cm 14cm 0cm 1.5cm},clip,width=160mm,scale=0.5]{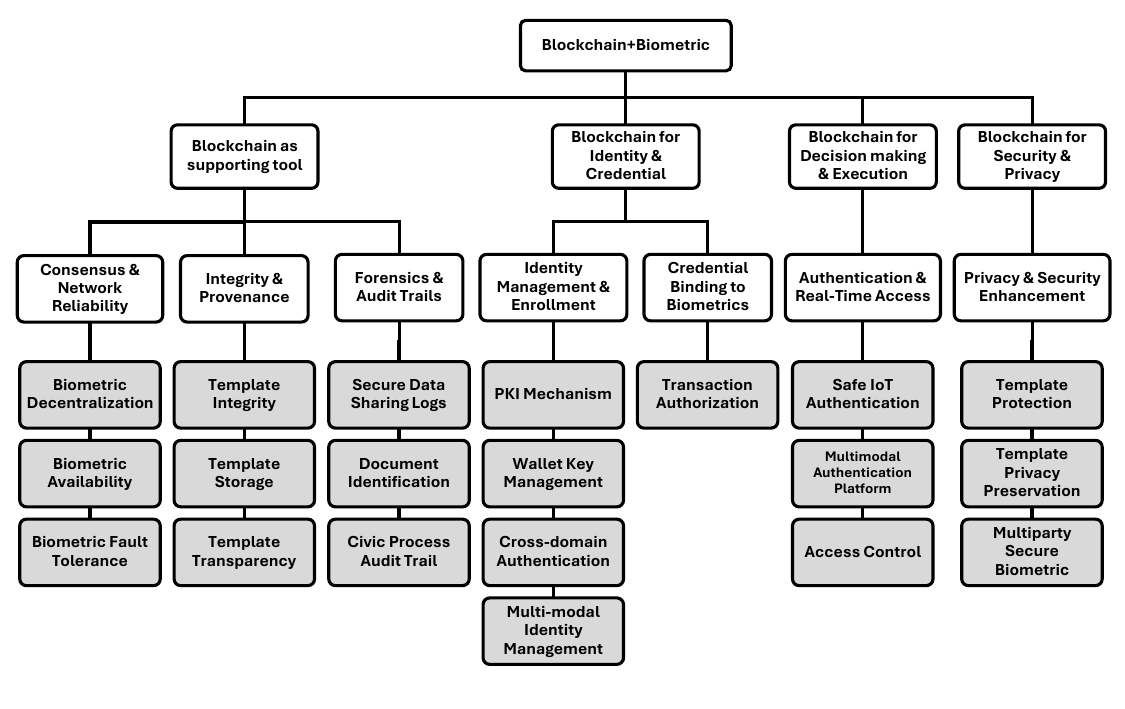}
 \caption{The Proposed Taxonomy of Current Blockchain–Biometric Integration Approaches}
 \label{fig:taxonomy}
\end{figure*}

Biometric systems can greatly benefit from the functionality and features offered by the blockchain. Fig.~\ref{fig:taxonomy} presents our proposed taxonomy of blockchain features applied to biometric integrations, based on existing works. In this section, we discuss the strengths and weaknesses of key aspects that play a pivotal role in the integration of blockchain and biometrics.



\subsection{Scalability}

\subsubsection{Storage}
All blockchain-based applications face an inherent challenge related to the limited block size and the continuous growth of stored data \cite{zhou2020solutions}. In such systems, every transaction that is processed through the network must be permanently recorded on the blockchain, and all participating nodes are responsible for maintaining the complete transaction history. With the current size and growing storage space of two prominent public blockchains, Bitcoin and Ethereum, exceeding 200GB \cite{Ref67}, scalability has become the main barrier to adopting the blockchain in all applications, including biometrics. Broadly, three categories of solutions have been proposed: 

The first solution is to remove old recorded transactions and keep the balance of non-empty addresses using a database called the account tree \cite{Ref91}. This approach obviates the need for storing all transactions by keeping the hash of previous transactions to check the validity of new transactions. 

The second solution involves redesigning blockchain protocols, such as Bitcoin-NG \cite{Ref92}. This approach divides conventional blocks into two parts: a key block and a microblock. Miners compete for the key block to become the next leader, who then generates microblocks to store new transactions. However, this design has limitations. For biometric verification, it must remain lightweight due to high usage rates, and storing all transactions on-chain raises storage concerns. Moreover, storing raw biometric data or embeddings directly on a public blockchain is legally risky, so typically only protected biometric templates are stored.


Lastly, the third solution, widely discussed in the recent literature \cite{Ref17, Ref22, Ref33, Ref35, Ref37, Ref38, Ref50, Ref54, Ref58, Ref59, Ref64, ghafourian2025nlml} is off-chain storage, which records biometric transactions outside the blockchain while maintaining cryptographic references on-chain. This approach, though it does not guarantee full immutability, offers the most practical scalability benefits. It is typically implemented either by (i) recording the hash of biometric data within the ledger while storing the data itself in an external repository, or (ii) recording only these hashes within a Merkle tree~\cite{Ref67}, which ensures integrity through a recursive process of self-construction.

A more advanced form of off-chain storage that preserves decentralization is the InterPlanetary File System (IPFS) \cite{Ref93}, a distributed file system connecting devices via a shared data structure. Like the blockchain itself, IPFS has no central authority or single point of failure. Most blockchain–biometric studies propose storing the hash of biometric data on-chain and the data itself on IPFS. However, this setup limits direct biometric matching, since hashes cannot be used for comparison. Biometric verification thus requires the protected template to remain accessible off-chain, where matching can be performed through smart contracts or auxiliary verification modules.

In summary, four main recording strategies exist for combining blockchain and biometrics: (i) storing biometric data directly on-chain, (ii) storing data off-chain on a side-chain with its hash on-chain, (iii) storing data in distributed file systems (e.g., IPFS) with on-chain hashes, and (iv) storing only hashes on a Merkle tree. The advantages and disadvantages of each approach are summarized in Table~\ref{tab:recApprochesComparison}.

\begin{table*}[tbh]
\centering
\caption{Current recording approaches for combining blockchain with biometrics found in literature \cite{Ref38, Ref68}: (i) Store biometric data as it is on-chain (on the public blockchain), (ii) Store biometric data off-chain on a side-chain (e.g., a private blockchain or any kind of side-chain) and the hash of it on-chain, (iii) Store biometric data in distributed file systems (e.g., IPFS) and the hash of it on-chain, (iv) Store only the hash of biometric data on a Merkle tree.}
\label{tab:recApprochesComparison}
\resizebox{16.5cm}{!}{%
\begin{tabularx}{1\textwidth}{>{\centering\arraybackslash}c>{\centering\arraybackslash}m{6.5cm}>{\centering\arraybackslash}m{7cm}}
\hline
\\ \textbf{Approach}  & \textbf{Advantages} & \textbf{Limitations}\\  \\ 
\hline
\hline
(i) 
  &
\begin{itemize}
  \item Can be used for almost all biometric applications
  \item Taking advantage of all characteristics of a public blockchain
  \end{itemize}
  &
  \begin{itemize}
  \item Storage space in the current public blockchain is expensive
  \item Slow (not suitable for real-time use-cases)
  \item	Wouldn’t comply with GDPR
\end{itemize}\\
\hline
(ii)
  &
\begin{itemize}
  \item Doesn’t have the scalability limitations of public blockchains (cheaper, faster)
  \end{itemize}
  &
  \begin{itemize}
  \item Since the data is stored off-chain, it's not possible to take advantage of public blockchain characteristics like immutability, transparency and decentralization
\end{itemize}\\
\hline
(iii)
  &
\begin{itemize}
  \item Doesn’t have the scalability limitations of public blockchains (much cheaper, faster)
  \item Decentralized networks like a public blockchain
  \end{itemize}
  &
  \begin{itemize}
  \item It’s a distributed file system, not a blockchain
  \item Like storing data in public blockchain it is subjected to legal issues
  \item Need two storage locations (public blockchain and IPFS)
  \item The hash of data is not useful for all biometric applications
\end{itemize}\\
\hline
(iv)
  &
\begin{itemize}
  \item Doesn’t have the scalability limitations of public blockchains (less storage space, faster)
  \end{itemize}
  &
  \begin{itemize}
  \item Very limited in terms of the biometric application
\end{itemize}\\
\hline

\hline
\end{tabularx} 
}
\end{table*}


\subsubsection{Latency}
Writing new blocks on-chain is quite slow and completely infeasible for being adopted in real-time applications. For example, in terms of the most popular cryptocurrency, the required time to establish a new block in the Bitcoin network is 10 minutes on average, making it limited to processing up to 7 transactions per second. The same scalability limit lies ahead of the second generation of blockchain systems such as Ethereum with a limit of 15 transactions per second. As a result, biometric applications that require real-time responses, such as most verification and identity management systems, cannot integrate with these blockchain networks. To address this issue, next-generation blockchain networks overcoming these limitations should be developed. Off-chain storage has also been proposed to mitigate the latency problem \cite{Ref9}. In this case, IPFS can retrieve the stored content more quickly and reliably by locating it through its file information. Although off-chain storage helps reduce latency compared to on-chain solutions, it still imposes limitations on certain biometric applications (e.g., biometric authentication, surveillance, e-commerce), particularly similar to the storage challenges previously mentioned. 

\subsubsection{Computation cost}
It is discussed that another reason to suggest using off-chain storage is attributable to the total cost of the computation on-chain. The main biometric operations that impose arithmetic costs on the blockchain network involve setup, template transformation, encryption, decryption, upload (read, write), updating the smart contract, and calculating distance. However, the exact computational cost of biometrics in the blockchain is highly dependent on the template acquisition method \cite{Ref35}, the underlying consensus protocol, and the operations executed by the contract \cite{Ref22}. For example, the initial cost for reading and writing 1KB of data on the Ethereum network is 6,400 and 640,000 gas units, respectively \cite{Ref68}. Therefore, storing 1 KB of data from the vector of face characteristics, for instance from the VGG-Face template, on the Ethereum network costs approximately 352,912 gas. Gas is a unit to pay for transactions on the Ethereum blockchain denoted in gwei, where 1 ETH = $1* 10^9$ (1,000,000,000) gwei. It is a measure to compute how much a transaction would cost. Therefore, at the time of writing this article, with a gas price of $0.635$ gwei per gas, the cost of storing 1 KB of face feature vector data on a full on-chain storage scheme is approximately \$1.03 (1 ETH = \$4,596.57). Due to the computationally intensive process of biometric applications such as biometric verification, the widely suggested solution for massive scalability appears to be to resort to off-chain storage (approaches ii and iii in Table \ref{tab:recApprochesComparison}), regardless of the limitations. 

\subsection{Smart Contracts}
Smart contracts can automatically execute verification processes when predefined biometric conditions are met, eliminating the need for intermediaries. In addition, the use of smart contracts on the blockchain enables users to grant or revoke permission for the usage of their biometric data. When implementing biometric applications on a blockchain using smart contracts, it is essential to consider two key issues: First, it is reported that gas estimation at the source code level is a difficulty for smart contract development. Since biometric applications are resource intensive, the underlying smart contract must be developed in a resource-friendly way. Second, special attention is needed for smart contract security and bugs, especially considering the immutability and non-reversibility of transactions. The record on the blockchain, the execution and the management of biometric templates will be handled by the smart contract. Therefore, any existing security problem or bug within the smart contract leads to a catastrophe. For example, more than 60 billion dollars were stolen from the DAO, Decentralized Autonomous Organizations that are entities to operate through smart contracts, due to the recursive call bug \cite{Ref97}. As an alternative solution to these problems, an off-chain smart contract has been proposed \cite{Ref98}. Unlike on-chain blockchains, they are executed outside the blockchain by only interested participants. Although on-chain protocols are better for low-cost computation and non-sensitive transactions, an off-chain smart contract is suggested for operations involving high-cost computation or private information of participants. However, it is better to exploit the advantage of an on-chain smart contract jointly for its auditability and trustable characteristics regarding any disputes about the execution of an off-chain protocol.

\subsection{Consensus}
In the blockchain context, consensus is a mechanism to reach an agreement between participants who do not necessarily trust each other in a distributed network without a central authority. The consensus protocol is responsible for the correct execution of the smart contract on the blockchain. In combination with biometrics, the consensus algorithm affects the latency to execute operations. With the emergence of the next generation of blockchain networks \cite{Ref9}, equipped with fast consensus algorithms that maintain the security benefits of their predecessors, the need for off-chain protocols, at least to overcome scalability limits, will likely be eliminated. To this end, the third generation of blockchain systems is emerging that targets scalability limitations. As an example of these new generations of consensus algorithms, Stellar consensus protocol \cite{losa2019stellar}, which is based on a new type of federated byzantine agreement \cite{2024_ICIP_SaFL}, can process 1,000 transactions/second, Ripple can handle up to 1,500 transactions/second, and Hashgraph, which takes advantage of a new type of asynchronous, byzantine fault tolerance can achieve above 250,000 transactions/second. Recently, an identity management method called DID (Distributed Identifiers) \cite{Ref86} has been introduced which transfers computation overload off-chain, decreasing demands for the main blockchain. Due to its decentralized nature and ability to enable trusted interactions, DID serves as a viable alternative for storing biometric templates on-chain. We will discuss DID further in Section \ref{ssec:Promising-usecases}.

\subsection{Security Enhancement}
Owing to its unique features, blockchain can potentially help improve the security of distributed networks. In terms of the combination with biometrics, while it offers a breakthrough, multiple concerns need to be addressed. 

First and foremost, this integration can not only help introduce a public key infrastructure (PKI) platform \cite{maurer1996modelling} to biometrics, but also helps to overcome the problem of a single point of failure with which conventional PKIs are entangled. In conventional PKI systems, if the Certificate Authority (CA) is not accessible, it is not possible to verify the authenticity of generated certificates. Due to its availability characteristics, blockchain addresses this shortcoming. Second, considering biometric verification, it provides a distributed storage system in which multiple transformations of the same biometric template can be stored in different places \cite{2022_WACVw_OTB-morph_MG,2023_MIR_OTB-morph_Mahdi}. In existing scenarios, some tried to increase blockchain security by generating biometric cryptographic keys to secure the private key in blockchain wallets \cite{Ref62} or creating an encryption scheme to sign blockchain transactions \cite{Ref57}, while others took advantage of the blockchain to increase biometric system security. 

Regarding the former, because the private key resides in the user’s wallet, its security depends on the strength of the wallet’s password. Biometrics can also be used as an access control to the wallet in a key binding approach \cite{Ref61}. As for the latter, our focus in this survey, most studies try to secure biometric verification by leveraging blockchain for decentralized auditability, availability, and immutability. Some studies suggest blockchain for biometric applications where transparency is required, such as public data management, electronic voting, and taxation \cite{Ref64, Ref50}. 

Since the blockchain is immutable and the recorded information persists there forever, the fate of stored biometric templates would be shifted to the security of the underlying cryptographic method. Concerning auditability, referring to the 51\% attack \cite{Ref95}, it is possible to exert influence by most of the nodes that colluded for a single purpose and contradict the facts already recorded by creating a fork. Furthermore, due to the immutability of the blockchain, it is not possible to discard an expired biometric template stored earlier on a public ledger unless $51\%$ of the nodes colluded. Regarding private blockchains where a single organization controls the validator set, it is practically possible to remove an expired biometric template by reverting the ledger to a prior state (or producing an agreed fork) and then producing new blocks from that point. However, this process is costly, since removing a biometric template from the blockchain also deletes all subsequently added templates. Thus, the owner of the private blockchain needs to rewrite the history after the discarded block. 

A recently proposed solution for Bitcoin introduces a miner-redactable and scalable approach that reportedly eliminates the need for trustees, heuristics, or hard forks. Kiraz et al.~\cite{kiraz2024redact} claim to achieve instant erasure of all transaction types by enforcing transaction ``templates''. These templates ensure that only the non-executable portions of unlocking scripts can be modified, preventing unintended alterations. In particular, a significant portion of such nonexecutable data consists of nonfinancial information such as biometric templates or references to them within the Bitcoin network, which are stored as OP\_RETURN data. In 2017, reports indicate that 86.8\% of Bitcoin’s non-financial data falls into this category. Furthermore, this approach eliminates the need for modifications in child transactions after redaction, effectively resolving the issue of ``cascade of changes'' found in $\mu$Chain~\cite{puddu2017mu}, a redactable blockchain designed for Ethereum.

\subsection{Privacy Preservation}
In general, each node in the network must execute and validate a smart contract on the chain. To this end, all relevant transactions are visible to the entire blockchain network, resulting in privacy leakage. Privacy leakage has been shown to occur on the blockchain even when users rely only on their public key to create a transaction \cite{Ref100}. Since the blockchain is transparent, it is possible to link transactions made using the same public key to an identity. When it comes to the processing of biometric data for very sensitive applications such as e-health, it is essential to address data privacy and anonymity issues \cite{ghafoorian2020anonymous}. In addition, since public keys are bound to biometric templates, linking a public key to a real identity may result in permanent leakage of the biometric data. Although homomorphic encryption can be used to keep and process biometrics in cipher for fixed length templates \cite{2016_CVPR_HEfixedLength_Marta,Ref7}, and other privacy-preserving methods can be applied for variable length templates \cite{2017_Access_HEmultiDTW_Marta}, there is no guarantee that the protected template will remain unbroken in the future. 

Bitcoin provides only weak identity privacy. To overcome this limitation, several privacy coins have been introduced, including Dash, Zerocoin, Monero, Zcash, Zether, Grin, Verge, and MimbleWimble, among others \cite{zhang2023privacy}. These cryptocurrencies aim to enhance transaction anonymity and protect user identities, making them relevant to privacy-focused, permissionless blockchains designed for identity privacy, which is crucial for biometric systems.

One of Bitcoin’s main privacy vulnerabilities is that data miners can link individual transactions and track user activities. A common solution to this issue is the use of Bitcoin laundering services, also known as mixers, which mix multiple users' bitcoins to obscure transaction histories. Hence, one of the most common recommendations is to employ a laundry service such as CoinShuffle++, which facilitates coin mixing to enhance privacy \cite{ruffing2014coinshuffle}.

Zerocoin (formerly Zcoin, XZC) is a proposal to extend Bitcoin with anonymous transactions. This extension would have functioned as a money laundering pool, where bitcoins were temporarily pooled in exchange for a temporary currency called zerocoins. Although laundering pools are already used by various currency laundering services, Zerocoin aimed to integrate this functionality at the protocol level, eliminating the need for trusted third parties. This Ensures anonymity using cryptographic principles while still recording transactions within the Bitcoin blockchain \cite{miers2013zerocoin}.

Monero (XMR) was the first blockchain designed with the privacy and resistance to censorship as the core principles. Every Monero transaction is anonymous by default, unlike Bitcoin, where transparency is the norm. Monero uses the CryptoNote protocol, which employs ring signatures and one-time keys to obscure the origins and destinations of the transactions. This approach ensures strong privacy, making Monero one of the most widely adopted privacy coins with a high market capitalization \cite{van2013cryptonote}.

Zcash (ZEC) is another major privacy coin that offers confidential transactions. Unlike Monero, where privacy is enforced by default, Zcash provides users with the option to choose between shielded (private) and transparent (public) transactions, granting greater control over privacy preferences \cite{Zcash(ZEC)}.

These privacy coins demonstrate how permissionless blockchains can be leveraged for identity privacy. By integrating cryptographic techniques such as ring signatures~\cite{rivest2001leak}, zero-knowledge proofs~\cite{goldwasser2019knowledge}, and coin-mixing mechanisms~\cite{bonneau2014mixcoin}, these blockchain solutions provide individuals with enhanced control over their biometric data, mitigating the risks associated with traceable transactions.
More sophisticated general privacy-preservation mechanisms such as differential privacy, to our knowledge, have not yet been adapted and studied to blockchain. 

Finally, biometric-specific privacy preservation methods (e.g., obscuring sensitive elements on biometric embeddings \cite{2021_TPAMI_SensitiveNets_Morales, 2023_WACV_Multi-IVE_Pietro}) can be applied directly.



\subsection{Literature Categorization}

Table~\ref{tab:techComparsions} provides a comparative overview of existing works that integrate blockchain technology with biometric systems, focusing on key technical aspects. The comparison highlights crucial parameters such as the biometric type (modality), the blockchain type (permissioned vs. permissionless), framework, storage approaches (on-chain vs. off-chain), integration purpose, blockchain network, and key technologies.

This comparison aims to identify the strengths and limitations of different approaches, providing insight into how blockchain enhances biometrics for the given integration purpose. By analyzing these technical attributes, researchers and practitioners can better understand the trade-offs involved in various implementations and identify potential areas for future improvement. Given that, only a few works include prototypes and even those lack standardized metrics (latency, throughput, storage cost), a direct quantitative comparison is not feasible. Accordingly, Table III presents a semi-quantitative aggregation, derived from the attributes cataloged in Table~\ref{tab:techComparsions}.

\def\lc{\left\lceil}   
\def\rc{\right\rceil}
\def\lf{\left\lfloor}   
\def\rf{\right\rfloor}

\begin{table*}
\renewcommand{\arraystretch}{2}
\footnotesize
\rowcolors{2}{gray!10}{white}
\centering
\caption{Comparison of existing works in terms of key technical information (N/S stands for Not Specified).}
\label{tab:techComparsions}
\resizebox{\textwidth}{!}{%


\begin{tabular}{ l P{2cm} P{1.5cm} P{2.5cm} P{2cm} P{3.5cm} P{1.5cm} P{3cm}}

\hline
\textbf{Scheme}      
& \textbf{Biometrics storage location}
& \textbf{Biometric modality}   
& \textbf{Framework}                      
& \textbf{Blockchain Type}
& \textbf{\textbf{Integration purpose}}                                         
& \textbf{Blockchain Network} 
& \textbf{Key technologies}  \\ 
\hline
\hline
\cite{Ref21}   
& On-chain  
& Fingerprint  
& Biometric authentication 
& Permissionless
& Safe storage of biometric vectors in blockchain 
& Ethereum 
& MetaMask, Cross-numbering  \\ 
\hline

\cite{Ref30}      
& On-chain
& Fingerprint 
& Biometric authentication 
& N/S 
& Availability, immutability, decentralization, anonymity
& N/S                                 
& N/S     \\ 
\hline

\cite{Ref36}   
& On-chain 
& Fingerprint 
& Biometric authentication 
& N/S 
& Providing availability           
& Ethereum                 
& N/S                 \\ 
\hline

\cite{sarier2022privacy}   
& On-chain 
& Fingerprint 
& Biometric authentication 
& Permissionless
& Privacy preservation       
& Monero                 
& Public key cryptography, Zero knowledge proofs               \\ 
\hline

\cite{Ref63}      
& On-chain  
& Fingerprint                         
& Electronic voting system  
& Permissionless 
& Cost efficiency, trust
& Bitcoin                  
& Elliptic Curve Digital Signature (ECDSA)           \\ 
\hline

\cite{Ref48}        
& On-chain 
& Fingerprint, face   
& Secure biometric recognition 
& Permissioned
& Combat tampering, provide fault tolerant  
& N/S                    
& 
Multi-PIE, MEDS-II, Public key cryptography \\ 
\hline

\cite{Ref47}        
& On-chain 
& Fingerprint, face
& Biometric identification   
& Permissionless  
& Providing privacy      
& Bitcoin                   
& Elliptic curve, RSA      \\ 
\hline

\cite{buchmann2017enhancing}
& On-chain
& Fingerprint, iris
& Identity management
& Permissionless
& Long-term security,  post quantum security
& Bitcoin
& directed acyclic graphs (DAG) \\
\hline

\cite{Ref23}
& On-chain 
& Fingerprint, finger vein  
& Multimodal biometric    
& N/S          
& Scalability, security
& N/S              
& N/S      
\\ 
\hline

\cite{Ref20}      
& On-chain  
& Finger vein  
& Biometric authentication  
& N/S 
& Data integrity and availability 
& N/S                      
& Steganography, Gaussian smoothing, segmentation
\\ 
\hline

\cite{Ref34}        
& On-chain 
& Fingernail 
& Fingernail analysis management system    
& N/S 
& Privacy, immutability and trust 
& N/S                         
& HOG, LBP          \\ 
\hline

\cite{hagui2024blockchain}        
& On-chain 
& Face 
& Biometric authentication    
& N/S 
& Secure communication 
& N/S                         
& Shift-AES         \\ 
\hline

\cite{Ref27}
& On-chain  
& Face 
& Biometric authentication  
& N/S
& Availability, immutability     
& N/S                     
& Fuzzy extractor and secure sketch    \\ 
\hline

\cite{chen2022cross}      
& On-chain 
& Face  
& Biometric authentication    
& Consortium blockchain  
& Cross-domain authentication
& N/S                       
& N/S              \\ 
\hline

\cite{Ref52}        
& On-chain   
& Face 
& Reliable food logistics 
& N/S   
& Security, scalability ad privacy 
& N/S                       
& N/S            \\ 
\hline

\cite{Ref53}      
& On-chain 
& Face  
& Secure vehicle data-sharing    
& N/S   
& Further security                   
& N/S                       
& N/S              \\ 
\hline

\cite{Ref68}        
& On-chain  
& Face, signature  
& Biometric templates storage    
& N/S       
& Availability of Biometric templates       
& Ethereum                 
& Ropsten Ethereum tesnet, Biosecure DS2   \\ 
\hline

\cite{Ref51}  
& On-chain  
& Signature 
& Biometric signature  
& N/S 
& N/S                       
& N/S                       
& N/S                 \\ 
\hline
\cite{Ref25}        

& On-chain  
& All modalities
& Authentication     
& Permissioned  
& Providing traceability and reducing fraud  
& Hyperledger fabric                             
& N/S             \\ 
\hline

\cite{Ref28}    
& On-chain  
& N/S 
& Biometric authentication   
& Permissioned 
& Decentralization, auditablity, biometric template security                    
& Ethereum                     
& Geth/Solidity  \\ 
\hline

\cite{Ref55}     
& On-chain 
& N/S 
& IoT safe access   
& N/S  
& Transparency                          
& N/S                        
& N/S          \\ 
\hline

\cite{Ref45}  
& On-chain 
& N/S   
& Biometric based signature
& N/S   
& Providing PKI mechanism in blockchain              
& N/S                      
& Fuzzy signature      \\ 
\hline

\cite{Ref54} 
& On-chain
& Finger vein  
& Library management    
& N/S  
& Availability          
& N/S            
& N/S                \\ 

\hline

\cite{Ref50}        
& On-chain  
& Fingerprint, iris   
& Electronic document identification  
& N/S          
& Traceability, security, scalability 
& N/S     
& N/S             \\ 

\hline

\cite{alzahab2024biometricidentity}        
& Off-chain and On-chain 
& N/S  
& biometric authentication 
& Permissionless         
& Decentralized authentication, IdM without re-enrollment
& Ethereum   
& Fuzzy Commitment Scheme (FCS)            \\ 

\hline

\cite{bao2023two}        
& Off-chain and On-chain 
& Fingerprint 
& Authentication
& Permissioned        
& Decentralized authentication
& Hyperledger Fabric   
& Fuzzy Extractor, Fabric chain           \\ 

\hline

\cite{Ref35}      
& Off-chain (IPFS) and on-chain  
& Fingerprint  
& Identity management   
& Permissionless     
& Providing Immutability to biometrics     
& Ethereum 
& IPFS   \\ 
\hline

\cite{Ref26}  
& Off-chain and~on-chain  
& Fingerprint, face   
& Biometric authentication   
& Permissionless   
& Transparency, convenience   
& N/S       
& Decision tree           \\ 
\hline

\cite{Ref17}     
& Off-chain:~on-device  
& Fingerprint   
& Biometric authentication   
& Permissionless
& N/S                         
& N/S           
& ORB descriptor  \\ 
\hline

\cite{Ref37, Ref38} 
& Off-chain 
& Fingerprint  
& Blockchain-based Identity management  
& Permissionless  
& Prevent credential transfer     
& Bitcoin                   
& Fuzzy extractor, fuzzy vault           \\ 
\hline

\cite{Ref33}  

& Off-chain:~on-device 
& Face 
& Identity management  
& N/S   
& Cost effective identity binding                
& N/S            
& TrustZone     \\ 
\hline


\end{tabular}
}
\end{table*}

\begin{table*}[t]
\renewcommand{\arraystretch}{2}
\footnotesize
\rowcolors{1}{gray!10}{white}
\centering
\resizebox{\textwidth}{!}{
\begin{tabular}{ l P{2.5cm} P{1.5cm} P{2.5cm} P{2cm} P{3.5cm} P{1cm} P{3.5cm}}

\hline

\cite{bathen2019selfis}  
& Off-chain: IBM’s Blockchain
& Face 
& Identity management  
& Permissioned   
& Security and privacy               
& Hyperledger Fabric           
& Bloom Filter    \\ 
\hline

\cite{Ref58, Ref59}    
& Off-chain:~wallet  
& Iris    
& Blockchain authorization 
& N/S  
& Addressing double spending      
& N/S           
& IIF algorithm, Spatial histogram    \\ 
\hline

\cite{Ref64}      
& Off-chain and~side-chain  
& Iris 
& Online voting system  
& Permissionless  
& Decentralize and tamper proof storage of biometric templates        
& Ethereum 
& Meta mask, Gabor filter, Rutovitz crossnumbering      \\ 
\hline

\cite{Ref22}        
& Off-chain:~IPFS 
& N/S 
& Biometric authentication   
& N/S
& Cost effective identity binding     
& N/S                 
& IPFS, homomorphic encryption     \\ 
\hline

\cite{Ref29}    
& N/S 
& Face 
& Authentication    
& Permissioned 
& N/S                         
& N/S                            
& Fuzzy vault     \\ 
\hline

\cite{mishra2021pseudo}    
& N/S 
& Face  
& Authentication     
& N/S 
& Privacy, Revocability                
& N/S                            
&  random distance method (RDM)   \\ 
\hline

\cite{Ref57}  
& N/S 
& Face 
& Transaction authorization 
& N/S           
& Addressing double spending problem in blockchain                          
& N/S                             
& HOG face detector, BNIF algorithm    \\ 
\hline

\cite{Ref61}    
& N/S 
& Face  
& Key management           
& N/S 
& Securing wallet keys                 
& N/S                            
& HOG,~Fuzzy vault          \\ 
\hline

\cite{Ref62}   
& N/S 
& Behavioral biometric  
& Key management  
& N/S
& Continuous authentication    
& N/S                            
& 12-layer CNN network based on keras  \\ 
\hline

\cite{Ref24}  
& N/S 
& Psychological traits 
& Multimodal biometric authentication  
& Permissioned  
& Liveliness detection before authentication   
& N/S                             
& SaaS    \\ 
\hline

\cite{Ref67} 
& N/S                                 
& N/S  
& Biometric template protection 
& N/S   
& Immutability, accountability, availability 
& N/S                                  
& N/S       \\
\hline
\hline
\end{tabular}
}
\end{table*}

\section{Blockchain and Biometrics: GDPR Elements}\label{sec:Legal_review}

Unlike centralized databases, blockchains enable selective data sharing that can mitigate identity theft and privacy risks \cite{Ref12, finck2018blockchain}. However, they raise several GDPR-related challenges, analyzed below across five key parameters: biometric data storage, accountability, immutability, anonymization, and data protection by design and default.


\subsection{Biometric Data Storage}
This parameter is related to the questions as to how and where the template is stored. Interdisciplinary research \cite{Ref69} in 2009 concluded that biometrics could improve privacy depending on the design and safeguards. In addition to the research analyzed above, blockchain surveys show that depending on the types of network and storage, these systems might have different levels of GDPR compatibility with the right to data protection \cite{Ref70}. These two storage criteria are often suggested as key determinants of GDPR compatibility level, whereby a helpful distinction can be made based on on-chain and off-chain storage. 
 
 \subsubsection{Off-chain storage}
Typically, central storage solutions are not considered privacy-friendly, as observed by the European Data Protection Supervisor \cite{Ref71}. Off-chain storage of biometric references can be decentralized and such strict decentralization methods are considered more privacy friendly \cite{Ref72} as the data controller cannot access the biometric templates and only receives the comparison result. However, data security is a crucial part of the data protection law, and this type of decentralization may also be susceptible to attacks. As explained in various reports, such as by the BSI, local storage of biometric data enabled by software, for example, rich OS of a mobile handset, cannot be regarded secure \cite{TR03159}. ENISA further observed that multiple biometric templates can be stored in Touch or FaceID and can belong to more than one person \cite{eIDAS}.


A common misunderstanding in the technical papers analyzed is that off-chain biometric data storage is, by default, GDPR-compliant. However, as explained, off-chain storage itself does not make the processing compliant or secure. This implies that even if the biometric data are stored off-chain, personal data linked to it, such as a user ID, public keys, and the content of the transactions, are usually still stored on-chain. In such cases, even if the public keys, e.g., hashed identifiers, change in every transaction, they still qualify as personal data under the GDPR, as the natural persons remain identifiable. Nevertheless, off-chain storage of personal data is generally recommended, a practice also supported by the recent guidelines from the EDPB \cite{EDPB2025}.

\subsubsection{On-chain storage} On-chain processing, that is, the processing "located, performed, run inside a blockchain system" \cite{ISO22739} of the special categories as such, while risky for the reasons we will discuss in detail later, is not per se against the GDPR, as long as the processing is lawful. For instance, in cases where the explicit consent of the data subject or another legal and legitimate ground under Article 9 is relied on, and the other GDPR requirements are met, this processing by itself would not infringe the GDPR. 

\subsection{Accountability}

As explained previously, one of the most critical problems identified in the legal literature is how to allocate responsibility for data processing activities in the face of the polycentric nature of blockchains (i.e., accountability) \cite{Ref1, Ref2, Ref3}.
Although private blockchains are less inclined to have such problems, the real benefits of blockchains, namely transparency, can be seen in public blockchains. That is why it is not surprising that identity management systems based on blockchain (and biometrics) are positioned to run their projects on public blockchains. This can be seen as a general conceptual issue with respect to public blockchains and requires more legal guidance and research. 
The EDPB Guidelines acknowledge that participants' roles vary based on the governance model. In permissionless blockchains, nodes can influence processing and key decisions, such as forking. When nodes act independently, the EDPB recommends forming a consortium or legal entity among nodes. This legal structure would serve as the controller for processing \cite{EDPB2025}.
 Hence, unless more legal clarity is established on the issue, processing via public blockchains is risky, as the entity responsible for GDPR compliance is unclear.

This general concern about blockchains is amplified when it comes to biometric data processing, as they are universally distinctive enough to identify people. Furthermore, biometric data are not chosen by individuals and cannot be reissued. Thus, biometric data are considered more sensitive and their combination with blockchain technology can result in long-term linkages between personal data and biometric IDs \cite{goodell2019decentralized}.

Riva \cite{Ref74} suggests that an accountable agency should be assigned top to bottom to eliminate the accountability issue. The decentralized nature and benefits of blockchains are severely undermined in this model. The French Data Protection Authority (CNIL) asserts that professional or commercial activity should be the main parameter when assessing controllership \cite{lyons2018blockchain}. Although this may be a practical solution, the jurisprudence of the European Court of Justice holds that the controller should be determined through a factual analysis, which means that a factual influence is necessary for an entity to qualify as a data controller \cite{JUDGMENTOFTHECOURT(SecondChamber)}. 

\begin{table*}
\renewcommand{\arraystretch}{2}
\footnotesize
\rowcolors{2}{gray!10}{white}
\centering
\caption{Semi-quantitative summary of existing blockchain–biometric integration approaches.}
\label{tab:techSummaries}
\resizebox{\textwidth}{!}{%


\begin{tabular}{ l P{1.5cm} P{3cm} P{2.5cm} P{2.5cm} P{2cm}}

\hline
\textbf{Approach Type}      
& \textbf{\# Papers}
& \textbf{Common Blockchain Type}   
& \textbf{Implementation Status}                      
& \textbf{GDPR Compliance Risk}
& \textbf{Scalability}  \\ 
\hline
\hline
Full On-chain Storage 
& $24$ 
& Bitcoin, Ethereum  
& Mostly conceptual; few demos
& High risk
& Poor  \\ 
\hline

\hline
Off-chain + Blockchain Anchor 
& $10$
& Ethereum (+ FCS), Hyperledger Fabric  
& Multiple prototypes (Fabric; ETH dApp)
& Medium risk
& Moderate–Good  \\ 
\hline

\hline
Hybrid (mixed) 
& $3$
& Sidechains, Hyperledger  
& Prototype level
& Medium risk
& Moderate  \\ 
\hline

\hline
SSI / Interoperable Identity
& $4$
& Hyperledger, Ethereum
& Emerging pilots
& Low risk
& Good  \\ 
\hline

\hline
Audit Trails / Logging
& $2$
& Mixed / not specified
& Conceptual
& Medium risk
& Good  \\ 
\hline

\hline
Privacy-Enhancing Approaches
& $1$
& Mixed / not specified
& Conceptual only
& Low risk
& Poor  \\ 
\hline

\hline
\end{tabular}
}
\vspace{1mm}\\ 
\centering
\begin{minipage}{0.95\textwidth}
\centering
\scriptsize
\textit{Notes:} “FCS” stands for the \textit{Fuzzy Commitment Scheme}, a cryptographic method for securely binding biometrics to random keys. 
“Blockchain Anchor” denotes architectures where only cryptographic references (e.g., hashes or commitments) are stored on-chain, while biometric data remain off-chain.
\end{minipage}
\end{table*}

\subsection{Immutability}
Immutability as the main attribute of blockchain
may compromise the privacy of data subjects, particularly where their biometric data are processed \cite{Ref75}. Article 17 GDPR gives effect to the demands of individuals (data subjects) to have their data deleted without undue delay. As we just stated, biometric traits are unchangeable and biometric templates are, in theory, irrevocable. Therefore, any blockchain design that envisions raw or hashed on-chain storage of biometric templates would exacerbate the data protection risks. However, the same article also defines the exceptions to the right that are important to keep in mind when experimenting with blockchains and personal data. These exceptions demonstrate that the right to be forgotten is not an absolute right, nor is the right to privacy and the right to the protection of personal data. These exceptions are as follows:

\begin{itemize}
 
 \item Exercising the right of freedom of expression and information. The right to freedom of expression and information is one of the rights that might conflict with the right to protection of personal data. 
\item Compliance with a legal obligation which requires processing by the Union or Member States law to which the controller is subject for the performance of a task carried out in public interest. For example, research \cite{Ref63, Ref64} proposing an electronic voting system can be considered in the scope of this exemption.
\item For the performance of a task in the exercise of official authority vested in the controller. This exception can be the case if the controller processes personal data on-chain to carry out a legal obligation, such as demonstrating compliance with the tax regulation.
\item Reasons of public interest in the area of public health (e.g., processing personal data on-chain to facilitate compliance with the COVID-19 restrictions at the borders or public activities).
\item Archiving purposes in the public interest, scientific or historical research purposes, or statistical purposes. This exception provides a broad freedom to scientific projects processing personal data on-chain. However, other exceptions must be taken into account when products are placed on the market. For blockchain-based personal data processing, this exemption can be problematic, as it might be very difficult to identify a certain phase when scientific research ends and the product is placed on the market.
\end{itemize}


\subsection{Anonymity}
 Under GDPR, encrypted personal data are considered pseudonymous in principle, not anonymous. Pseudonymity refers to the cases where personal data are linked to unique identifiers, the processing of which is still subject to the GDPR. Article 29 Working Party states that 'removing directly identifying elements in itself is not enough to ensure that identification of the data subject is no longer possible'. It will often be necessary to take additional measures to prevent identification, again depending on the context and purposes of the processing for which the anonymized data are intended' \cite{ghafourian2023toward,Ref76}. However, most of the blockchain research analyzed uses the term anonymous to refer to data that are merely hashed or encrypted. This approach might create an incorrect understanding and implementation of the GDPR, where the responsible entities consider that the processing operation is not within the scope of the GDPR. It is specifically risky when it comes to hashed biometric templates, since these data are considered personal data \cite{Ref77}.

\subsection{Data Protection by Design and by Default}
For all the four problems explained above, another very relevant principle is data protection by design and default provided by Article 25 (1) GDPR. 
This principle is highly inclusive and starts to apply when the data controller determines the means of processing. Arguably, the developers of a blockchain-based biometric system can also be responsible. However, there are many arguments against this idea \cite{Ref73, Ref78}. The principle nevertheless requires the data controller to take into account the state of the art, the cost of implementation, the nature, scope, purposes of personal data processing, and the risks and severity for the rights and freedoms of the data subjects. Biometric data processing is by nature considered risky.
However, as mentioned, blockchain is still emerging and thus is considered a new technology whose impact on fundamental rights is unknown. Therefore, the processing of biometric data via blockchain is considered an innovative use of a large-scale operation that involves sensitive data where several datasets are matched or combined, and shall be subject to a Data Protection Impact Assessment (DPIA), under Article 35 GDPR\cite{Ref76}.



A common misunderstanding in this regard is that scientific papers use the term privacy, which refers solely to the security of the data. Privacy also encompasses informational privacy, that is, data protection.  As explained, data protection requires an array of technical and organizational measures, such as the principle of security that requires personal data to be kept confidential. This further means that personal data should be protected against accidental, unauthorized, or unlawful access, use, modification, disclosure, loss, destruction, or damage, according to Recital 39 of the GDPR. For example, using encryption techniques to pseudonymize personal data is considered an important element in data protection by design and default; however, other provisions of the GDPR should be respected when carrying out such processing, as pseudonymization itself does not render the processing GDPR-compliant.


\section{Future Directions and Use Cases}\label{sec:rec}

\subsection{Blockchain and Biometrics: Future Directions}
According to the results of previous research discussed in previous sections, we provide some suggestions and recommendations for future directions toward the combination of blockchain and biometrics that should be taken into account when designing a technolegal framework that incorporates blockchain in biometrics.

\subsubsection{Complete Biometric data should not be stored on a public ledger}
As discussed in section \ref{sec:Legal_review}, before having clarified the legal uncertainty on accountability, it is currently advised not
to store the complete biometric data on public blockchains.
A public blockchain is an immutable storage that is accessible to anyone \cite{ISO22739}. This means that once the data are published on the blockchain, it remains there forever. Considering the fact that biometric data contain information about our permanent identity and they are irreplaceable in case they leak. There are no guarantees that encryption mechanisms or biometric template protection methods remain completely safe in the future. 

From an efficiency standpoint, blockchain storage is currently computationally and economically expensive. In particular, reference \cite{Ref68} concluded practically that storing biometric templates on-chain or as hashed data is not cost-efficient. In addition to that, references \cite{Ref22} and \cite{Ref35} resorted to IPFS to solve the efficiency problem. Although using a distributed filesystem offers availability, it does not provide the security benefits that a blockchain provides, such as trust. 

Currently, approaches (ii) and (iii) in Table~\ref{tab:recApprochesComparison} are proposed to mitigate these concerns by keeping biometric data off-chain while anchoring proofs on-chain. However, a potential line of future research is to investigate GDPR-compliant designs that allow limited or partial on-chain storage of biometric data.

\subsubsection{Current integration is not suitable for real-time scenarios}
Currently, the process of publishing a new block and its confirmation time on a public blockchain is very slow. The latency of existing blockchain networks rules out real-time applications like biometric authentication and online payment. The most popular and widespread blockchain network, Bitcoin, is limited to publishing 3-4 transactions per second. At the time of writing this article, the time required for a Bitcoin blockchain network to issue a new block is 10 minutes. And this is without considering the confirmation time, which is variable and depends on the status of network. The second most popular public ledger, Ethereum, which is an open source and practical framework of most research implementations, is limited to 15 transactions per second with an average block publication time of 15 seconds \cite{Ref9}.

A promising research direction is to experiment with new-generation blockchain networks that overcome these latency and throughput limitations, such as Solana, which offer higher transaction speeds and lower confirmation times suitable for near real-time biometric integration.

\subsubsection{The integration should not be used for storing substantial data}
At present, blockchain storage is expensive and is not cost-effective for being used in applications that work with large volumes of data. Since all full nodes keep a copy of all transactions, keeping large amounts of data on-chain results in increased storage costs and processing expenses. Therefore, the scalability limit of the blockchain should be taken into account before using it as permanent data storage. 

\subsubsection{The integration should have economic justification}
The lifecycle of a permissionless blockchain depends on the miners who maintain the network. It should be economically interesting for these nodes to carry out their maintenance. To this end, the network pays a reward as a fee to miners who allocate their computing resources to maintain the network. For example, in Ethereum, this fee, known as gas, varies depending on the computational complexity of the operations executed within the network \cite{Ref67}. Therefore, running smart contracts on future frameworks that combine blockchain and biometrics should be economically justifiable.

\subsubsection{The integration should not rely on blockchain for privacy}
One of the intrinsic characteristics of blockchain is providing pseudonymity. This feature enables entities to exchange data with each other without knowing the identity of each other. As a result, some existing integration \cite{Ref30, Ref34} relies on this attribute to offer protection of privacy in biometric systems. However, blockchain entities are not fully anonymous, as they need to use a public key to send and receive transactions. In fact, such secrecy can be provided by encryption without the need to use a blockchain. In contrast, since the permissionless blockchain is transparent and the history of all transactions is accessible to anyone, users can be deanonymized using profiling \cite{Ref80}. Therefore, any framework that incorporates blockchain into biometrics should not consider blockchain as a privacy-preserving add-on. 

\subsection{Combining Blockchain and Biometrics: Use Cases}\label{ssec:Promising-usecases}
So far, the most attractive applications for integrating blockchain and biometrics have been biometric verification and identification. Although these are conventional and well-established biometric tasks, they are not necessarily the most promising use cases for this integration. After delving into the existing literature, we observed that in many cases, blockchain was not a fundamental requirement but rather incorporated due to its recent popularity. For instance, reference \cite{Ref52} itself notes uncertainty about why blockchain would be preferable to a conventional storage system when integrated with biometrics, yet still employs it. In the following, we propose several highly potential use cases where blockchain integration with biometrics could offer genuine technical or operational advantages, taking into account the strengths and limitations discussed throughout this paper.

\subsubsection{Using blockchain as computing power for biometrics}
The first consensus protocol employed in the blockchain network is Proof-of-Work (PoW), used in the Bitcoin network \cite{Ref81}. PoW is a complicated and resource-intensive strategy to select the miner who wins the competition, being entitled to publish the next block on the network. The difficulty level of this competition is continuously adjusted by the network to maintain trust and preserve the immutability of the blockchain. Despite these advantages, PoW is not economically efficient or environmentally friendly as it consumes substantial electrical power. When a block is published, only one node (miner) is rewarded for allocating its computational power, while the resources of other nodes who lose the competition are wasted. To remedy this issue, Proof-of-Stake (PoS) has been proposed \cite{Ref82}. However, PoS also has its problems regarding security threats and stakes issues \cite{Ref97}. A potential use case that benefits both blockchain and biometrics concerning PoW might be to leverage or rent out the computational power of miners. Biometric recognition can be a computationally expensive task, especially in relation to the model training phase. It is highly profitable to harness the computational power of miners in PoW-based blockchains to carry the burden of biometric computation for tasks that existed in different stages of a biometric system, such as training a biometric model, generating a protected template, and comparing biometric templates. In particular, it seems that the blockchain could be a profitable platform for both clients and model owners to operate a federated learning system \cite{aggarwal2021fedface} for biometrics. Therefore, in the proposed consensus algorithm, those nodes who dedicate their computational power to the benefits of the blockchain service for the biometric system are rewarded according to their contribution.

\subsubsection{Using blockchain as a platform for distributed identity management} 
\begin{figure}[tbp]
 \centering 
 \includegraphics[trim={5.8cm 1.5cm 7cm 1cm},clip,width=80mm,scale=0.5]{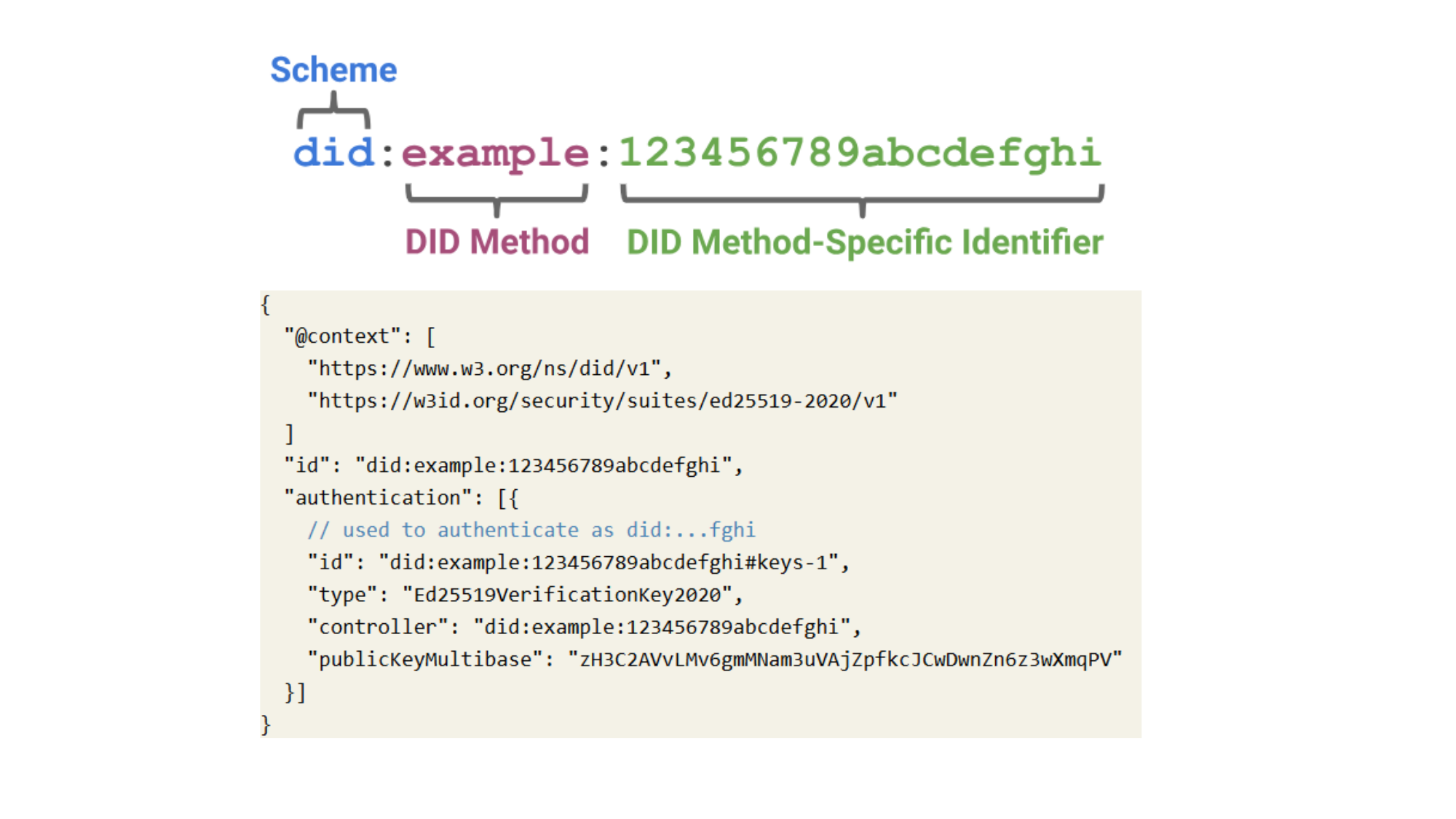}
 \caption{A simple example of a decentralized identifier (DID) and DID document.}
 \label{fig:DID}
\end{figure}

With the rapid growth pace of the digital world, it is very important to ensure appropriate access to resources available by our different identifiers (e.g., phone number, email address, domain names) across heterogeneous technology environments and platforms. Recently, an architecture called Decentralized Identifiers (DIDs)\footnote{https://www.w3.org/TR/did-1.0/} has been introduced as a new type of identity management system that allows for verifiable decentralized digital identity. According to the authors' definition, DIDs are URIs that associate a DID subject with a DID document, allowing trustable interactions associated with that subject. They have been designed to be decoupled from centralized registries, certificate providers, and identity suppliers. Figure~\ref{fig:DID} shows an example of DID.

This structure benefits the biometric system by fully controlling the subject's identifiers with an individual biometric identifier and independently of different heterogeneous providers. Thus, biometrics could significantly improve the currently distributed digital identity schemes based on blockchain \cite{Ref67}. In terms of on-chain challenges, DIDs help to transfer most of the processing load from the main blockchain to the off-chain network. A DID document consists of three fundamental parts: an authentication key, an authentication method, and a service endpoint. Whoever has access to the key is entitled to access the DID document. An example of this use case is the work of \cite{mishra2021pseudo}, which proposed a blockchain-based biometric authentication system where biometric data is decentralized and managed using DID and DID documents. We recommend that future research use biometric data only in combination with similar specifications, such as off-chain storage, to minimize both the quantity and sensitivity of personal data stored on-chain. 

\subsubsection{Using blockchain as a foundation for biometrics-based PKI mechanisms}

Public Key Infrastructure (PKI) comprises the policies, processes, server platforms, software, and workstations necessary to administer public-private key pairs and digital certificates. Its primary functions include issuing, maintaining, and revoking certificates that associate a public key with a specific entity, enabling secure communication and authentication within network environments\cite{Kuhn2001}.

Integrating biometrics with PKI is an emerging approach aimed at enhancing security and user authentication. However, the adoption of biometric-based PKI systems is not yet widespread. Traditional PKI relies on passwords or hardware tokens for authentication and is managed by centralized authorities, making it vulnerable to single points of failure and inefficiencies. Incorporating biometrics can improve security by linking cryptographic keys to unique physiological traits. In contrast, blockchain-based PKI decentralizes key management in a network of nodes, reducing dependence on a central authority and improving resilience\cite{Brunner2020}. Unlike existing biometric-PKI systems that rely on a CA often alongside Registration and Validation Authorities (RA/VA) to issue and manage certificates, blockchain-anchored biometric PKI decentralizes trust and certificate-status publication. In conventional designs (e.g., BioPKI~\cite{BioPKI}), biometrics derive or protect key material (e.g., deriving a biometric public key and reconstructing the private key from the user’s biometric), while enrollment, issuance, and verification remain CA/RA/VA-controlled, creating single points of failure and limited transparency. A blockchain-anchored architecture reduces reliance on any single CA by recording integrity-protected bindings (hashes/commitments) between identifiers, keys, and biometric-derived artifacts on a distributed ledger, enabling tamper-evident audit trails, transparent revocation, and interoperable identity proofs through smart contracts or DIDs.

Despite this potential, several factors contribute to its limited adoption.

\begin{itemize}
    \item Cost and complexity: Implementing biometric authentication within PKI systems requires specialized hardware and software to capture and process biometric data, leading to increased costs and complexity compared to traditional methods\cite{Kuhn2001}. 
    \item Lack of trust: PKI has limitations in establishing trust due to concerns about relying solely on keys to represent identities \cite{Ref84}.
    \item Certificate's life-cycle management: The handling of expired certificates is a major problem that exists in all PKI mechanisms. 
    \item Standardization: The lack of universally accepted standards for integrating biometrics with PKI poses challenges to interoperability across different systems and platforms.

\end{itemize}

Blockchain's decentralized nature reduces the need for centralized authorities, potentially lowering infrastructure costs and simplifying system management. It also ensures that all participants, regardless of their location or system, follow the same protocols that facilitate the adoption of common standards across different platforms and systems.

\begin{figure}[hbtp]
 \centering 
 \includegraphics[trim={6.8cm 3cm 7cm 1.5cm},clip,width=45mm,scale=0.5]{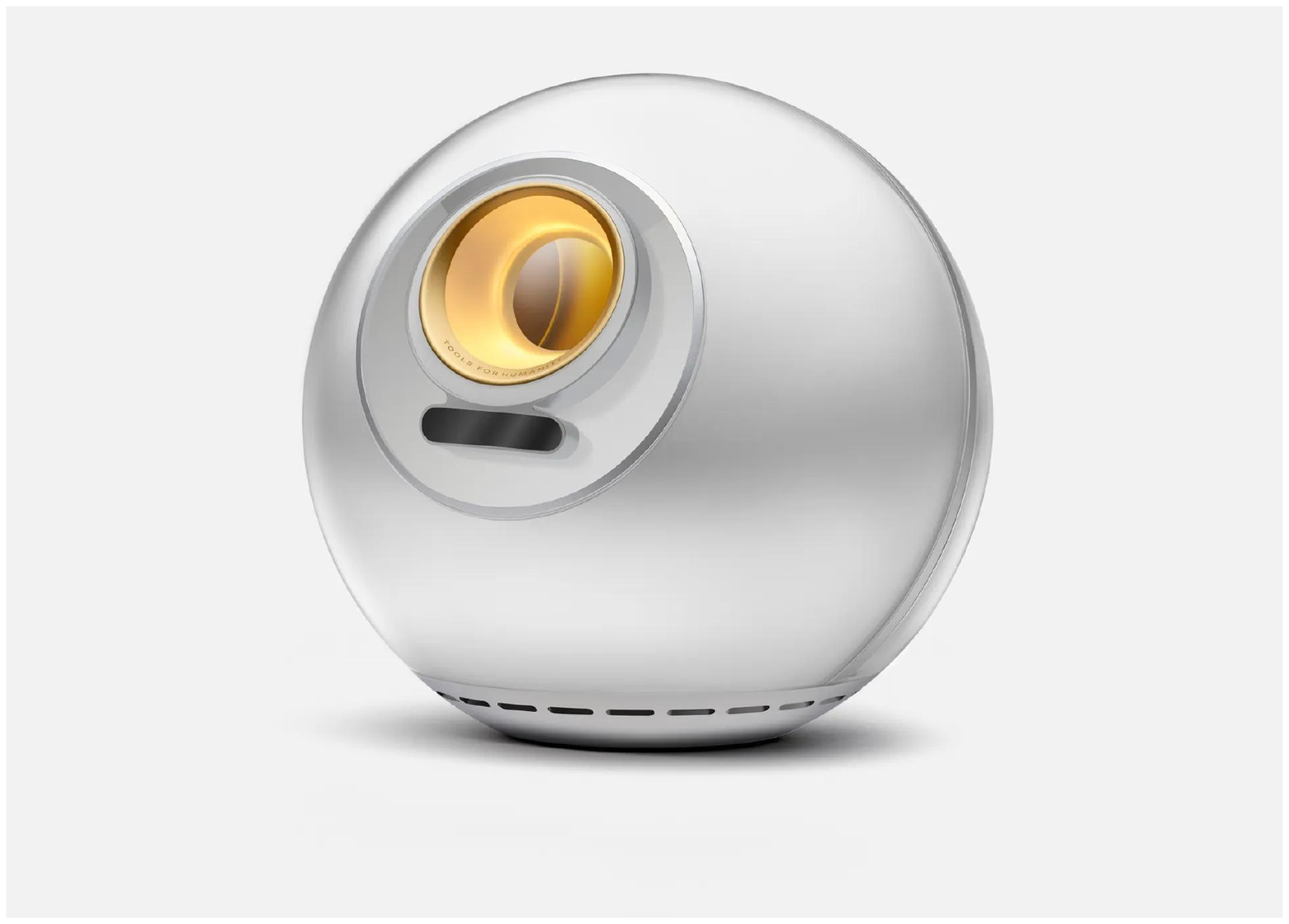}
 \caption{Iris-scanning Orb of the identity-verification project Worldcoin, now called World~\cite{wired2023worldcoin}.}
 \label{fig:Orb}
\end{figure}

World Network (formerly Worldcoin) and Humanity Protocol are other examples of biometric-based PKI systems that use blockchain technology. World Network employs the biometric iris scan depicted in Fig.~\ref{fig:Orb} to verify human identity, creating private crypto tokens \cite{Freire-Santos2006,Freire2007_ICB} for applications such as universal basic income and fraud prevention in virtual communications\cite{wired2023worldcoin}. (Note that iris biometrics can also be captured with standard smartphone cameras \cite{2019_HBookSelfie_SuperSelfieFaceIris_Alonso}, and iris biometrics, like other biometrics, are also susceptible to potential attacks and vulnerabilities \cite{2023_Book-PAD_Iris_AM}.) Similarly, the Humanity Protocol uses palm scans \cite{wang14regionalFusionPalmprint} to authenticate online accounts, addressing issues such as bots \cite{2021_EAAI_BeCAPTCHA_Acien,2022_PR_BeCAPTCHA,2023_CVPRW_BeC-Type}, fake accounts, and online fraud, using blockchain and biometrics for secure identity verification\cite{reuters2025humanityprotocol}.

\subsubsection{Using blockchain as a multimodal biometric authentication system} 
Blockchain helps the multimodal biometric system by providing a decentralized, secure, and privacy-preserving mechanism to manage access to the distributed stored encrypted biometric templates. It addresses the vulnerabilities of centralized systems by eliminating single points of failure, reducing the risk of data breaches, improving system availability, and empowering users with control over their biometric data through smart contracts. The results of each modality can be published and linked on the blockchain. As a result, several applications can authenticate the same individual simultaneously. In addition, most of the time, the underlying biometric hardware that captures user biometrics is not made or maintained by the same party. Therefore, publishing the authentication result for each modality on the blockchain helps to eliminate inconsistencies between the underlying infrastructure of each hardware. These results can be linked using an identifier, for example, using DIDs. Therefore, it is possible to link the biometric modalities of individuals using DIDs on the blockchain. Although standardization and risk mitigation efforts, such as DIDs, are appreciated, it should be kept in mind that DID documents can still be linked to individuals, and hence they are considered personal data under legal rules like the GDPR. Another practice is the distributed storage with IPFS. To this end, a recent study~\cite{sharma2024multimodal} stored decentralized fuzzy vaults of facial biometric images and dorsal hand images on IPFS and their address on the Ethereum Virtual Machine (EVM)-based blockchain network. Once the addresses of the fuzzy vaults are recorded on the blockchain, they become immutable, making it extremely difficult for malicious attackers to manipulate or delete this crucial information without consensus from the network participants.

\subsection{Suggestions to Mitigate Legal Risks}
First of all, we suggest that complete biometric data and any other personal data, such as name, nationality, etc., should not be stored on-chain until further research confirms that protective measures are sufficient~\cite{wang2020self, lyons2018blockchain}.

Furthermore, to mitigate data protection risks, the objective of processing operations must be clearly stated and compared to the applicable provisions outlined above to comply with GDPR. For instance, in cases where the purpose of the processing is in the public interest (e.g., electronic voting system, public health, or processing personal data on-chain to facilitate compliance with the COVID-19 restrictions at the borders), the exemptions to the right to erasure (Article 17 GDPR) could be relied on.

When it comes to research purposes, the GDPR provides broad freedom to scientific projects processing personal data on-chain. However, this exemption cannot be relied upon when products are placed on the market. Hence, we suggest that the objective of processing operations must always be clearly stated and compared to the applicable provisions outlined above to comply with GDPR. This is not a simple procedure because blockchains operate in numerous industries and stakeholders. Furthermore, blockchains can serve various functions, from smart cities to medical records storage, meaning that many stakeholders should take the necessary technical and organizational measures to comply with GDPR. This multistakeholder framework not only challenges the right to be forgotten, but also points to a meeting point for many purposes of different sectors. However, when we look at the reasons for incorporation in several studies analyzed \cite{Ref26, Ref30, Ref34, Ref35, Ref64} we see that the immutability of the blockchain is the purpose of integration. In our view, processing on blockchains requires an answer to erasure requests of data subjects. The right to erasure is not likely to be ensured, as the data registered in the blocks are very difficult and costly to delete, even if a court decision orders such, e.g., in the case of unlawfully processed data. 

Finally, with respect to public blockchains, there is uncertainty as to which entity is accountable, that is, the data controller. In these regards, private blockchains seem to pose fewer issues, but do not provide the real benefits of decentralization.

\section{Conclusions}\label{sec:con}
This survey examined the integration of blockchain and biometric recognition systems, highlighting both their potentials and limitations. 
From a technical standpoint, blockchain offers a robust foundation for enhancing the integrity, auditability, and interoperability of biometric systems. Its decentralized nature enables trust without reliance on a central authority, and recent frameworks demonstrate its capacity to support secure identity management and transparent verification. While real-time biometric applications and on-chain storage of sensitive data remain challenging, notable progress has been achieved through off-chain storage with on-chain proofs, supported by distributed file systems such as IPFS and permissioned frameworks like Hyperledger. Moreover, the emergence of high-throughput blockchain networks holds great promise for overcoming latency and scalability constraints, paving the way for more responsive and efficient biometric systems.
From a legal point of view, one shall take into account the sensitive nature of biometric data and the specific emerging risks out of this combination, such as accountability. It is important when experimenting
with biometric data and blockchain, researchers keep both raw biometric data (images) and templates off-
chain. Moreover, the purpose of the processing should be clarified for each granular processing operation, and
the risks should be assessed in order to apply additional safeguards for a future-proof data protection design.
Better precision of the purpose of the processing can also help understand whether the exemptions to the right to
be forgotten (Article 17 GDPR) apply to that processing.
Finally, based on the reviewed literature and our analysis, we highlight several recommendations and research directions for future studies, including the development of GDPR-compliant on-chain designs, the exploration of high-throughput blockchain platforms for near real-time verification, and the investigation of blockchain as a decentralized infrastructure for biometric-based PKI.

\section*{Acknowledgment}
This work has been supported by projects: PRIMA (ITN-2019-860315), TRESPASS-ETN (ITN-2019-860813), INTER-ACTION (PID2021-126521OB-I00 MICINN/FEDER) and BBforTAI (PID2021-127641OB-I00 MICINN/FEDER).

\ifCLASSOPTIONcaptionsoff
  \newpage
\fi



%




\bibliographystyle{IEEEtran}
\bibliography{bibtex/bib/IEEEabrv.bib,bibtex/bib/IEEEexample.bib}

%

\begin{IEEEbiography}
 [{\includegraphics[width=1in,height=1.25in,clip,keepaspectratio]{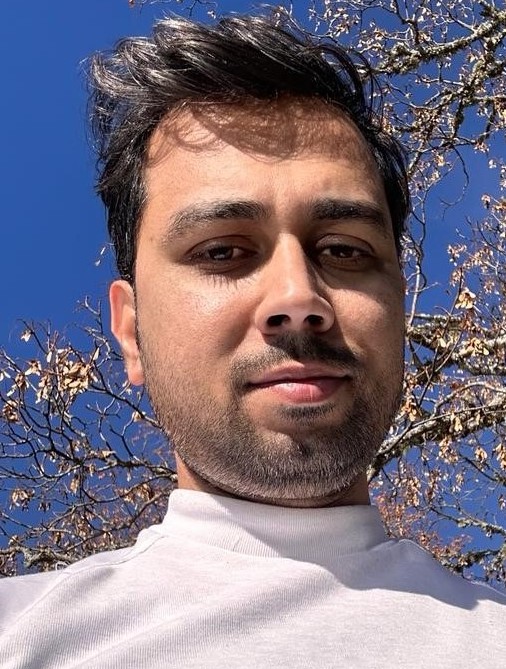}}]{Mahdi Ghafourian} received his B.Sc. degree in Computer Science from Islamic Azad University of Mashhad, Iran, in 2011, and his M.Sc. degree in Information Security and Assurance from Imam Reza University, Mashhad, Iran, in 2016, where he graduated second in his cohort in the Department of Computer Engineering and Information Technology. He completed his Ph.D. with Cum Laude distinction in 2024 at the Universidad Autónoma de Madrid, Spain, supported by a Marie Skłodowska-Curie scholarship under the PriMa (Privacy Matters) EU ITN project. Since 2024, he has been a researcher at the Intelligent Multimodal Vision Analysis (IMVA) group at Universitat Pompeu Fabra (UPF), Barcelona, working on the EMERALD project with a focus on real-time head pose estimation for AR/VR systems. His research interests include biometrics, adversarial robustness, generative models, and privacy-preserving machine learning.
\end{IEEEbiography}
\vskip -2\baselineskip plus -1fil
\begin{IEEEbiography}
[{\includegraphics[width=1in,height=1.25in,clip,keepaspectratio]{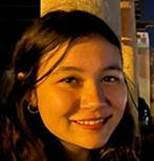}}]
{Bilgesu Sumer} is a doctoral researcher at KU Leuven (CiTiP), Biometric Law Lab (BLL). Her Ph.D. research focuses on the data protection implications of new digital identity systems with biometrics and blockchains. She is also interested in the regulation of AI,  data protection in transborder data flows, government access to personal data, certification, and liability.
\end{IEEEbiography}
\vskip -2\baselineskip plus -1fil
\begin{IEEEbiography}[{\includegraphics[width=1.in,height=1.25in,clip,keepaspectratio]{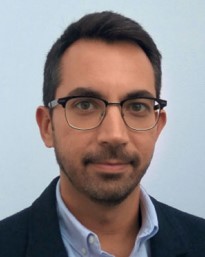}}]{Ruben Vera-Rodriguez} received the M.Sc. degree in telecommunications engineering from Universidad de Sevilla,
Spain, in 2006, and the Ph.D. degree in electrical and electronic engineering from Swansea University, U.K., in 2010. Since 2010, he has
been affiliated with the Biometric Recognition
Group, Universidad Autonoma de Madrid, Spain,
where he is currently an Associate Professor
since 2018. His research interests include signal
and image processing, pattern recognition, HCI,
and biometrics, with emphasis on signature, face, gait verification and
forensic applications of biometrics. Ruben has published over 100 scientific articles published in international journals and conferences. He
is actively involved in several National and European projects focused
on biometrics. Ruben has been Program Chair for the IEEE 51st International Carnahan Conference on Security and Technology (ICCST) in
2017; the 23rd Iberoamerican Congress on Pattern Recognition (CIARP
2018) in 2018; and the International Conference on Biometric Engineering and Applications (ICBEA 2019) in 2019.
\end{IEEEbiography}
\vskip -2\baselineskip plus -1fil
\begin{IEEEbiography}[{\includegraphics[width=1in,height=1.25in,clip,keepaspectratio]{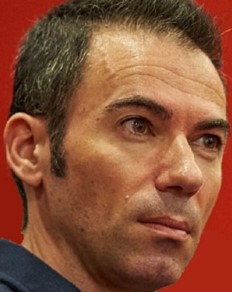}}]{Julian Fierrez} received the
The M.Sc. and Ph.D. degrees were obtained from Universidad
Politecnica de Madrid, Spain, in 2001 and 2006,
respectively. Since 2004 he has been at the Universidad Autonoma de Madrid, where he is Full Professor since 2022. His research focuses on signal and
image processing, AI fundamentals and applications, HCI, forensics, and biometrics for security
and human behavior analysis. He is an Associate
Editor for Information Fusion, IEEE Trans. on
Information Forensics and Security, and IEEE
Trans. on Image Processing. He has received best papers awards at
AVBPA, ICB, IJCB, ICPR, ICPRS, and Pattern Recognition Letters;
and several research distinctions, including EBF European Biometric
Industry Award 2006, EURASIP Best PhD Award 2012, Miguel Catalan
Award to the Best Researcher under 40 in the Community of Madrid
in the general area of Science and Technology, and the IAPR Young
Biometrics Investigator Award 2017. Since 2020 he has been a member of the
ELLIS Society.
\end{IEEEbiography}
\vskip -2\baselineskip plus -1fil
\begin{IEEEbiography}[{\includegraphics[width=1.in,height=1.25in,clip,keepaspectratio]{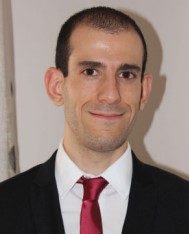}}]{Ruben Tolosana}
received the M.Sc. degree in
Telecommunication Engineering and his Ph.D.
degree in Computer and Telecommunication
Engineering, from Universidad Autonoma de
Madrid, in 2014 and 2019, respectively. In 2014, he
joined the Biometrics and Data Pattern Analytics -
BiDA Lab at the Universidad Autonoma de Madrid,
where he is currently working as a postdoctoral researcher. Since then, he has received
several awards such as the FPU research fellowship
from Spanish MECD (2015) and the European
Biometrics Industry Award (2018). His research interests are mainly focused
on signal and image processing, pattern recognition, and machine learning,
particularly in the areas of DeepFakes, HCI, and Biometrics. He is the author
of several publications and also collaborates as a reviewer in high-impact
conferences (WACV, ICPR, ICDAR, IJCB, etc.) and journals (IEEE TPAMI,
TCYB, TIFS, TIP, ACM CSUR, etc.). Finally, he is also actively involved in
several National and European projects.
\end{IEEEbiography}
\vskip -2\baselineskip plus -1fil
\begin{IEEEbiography}[{\includegraphics[width=1.in,height=1.25in,clip,keepaspectratio]{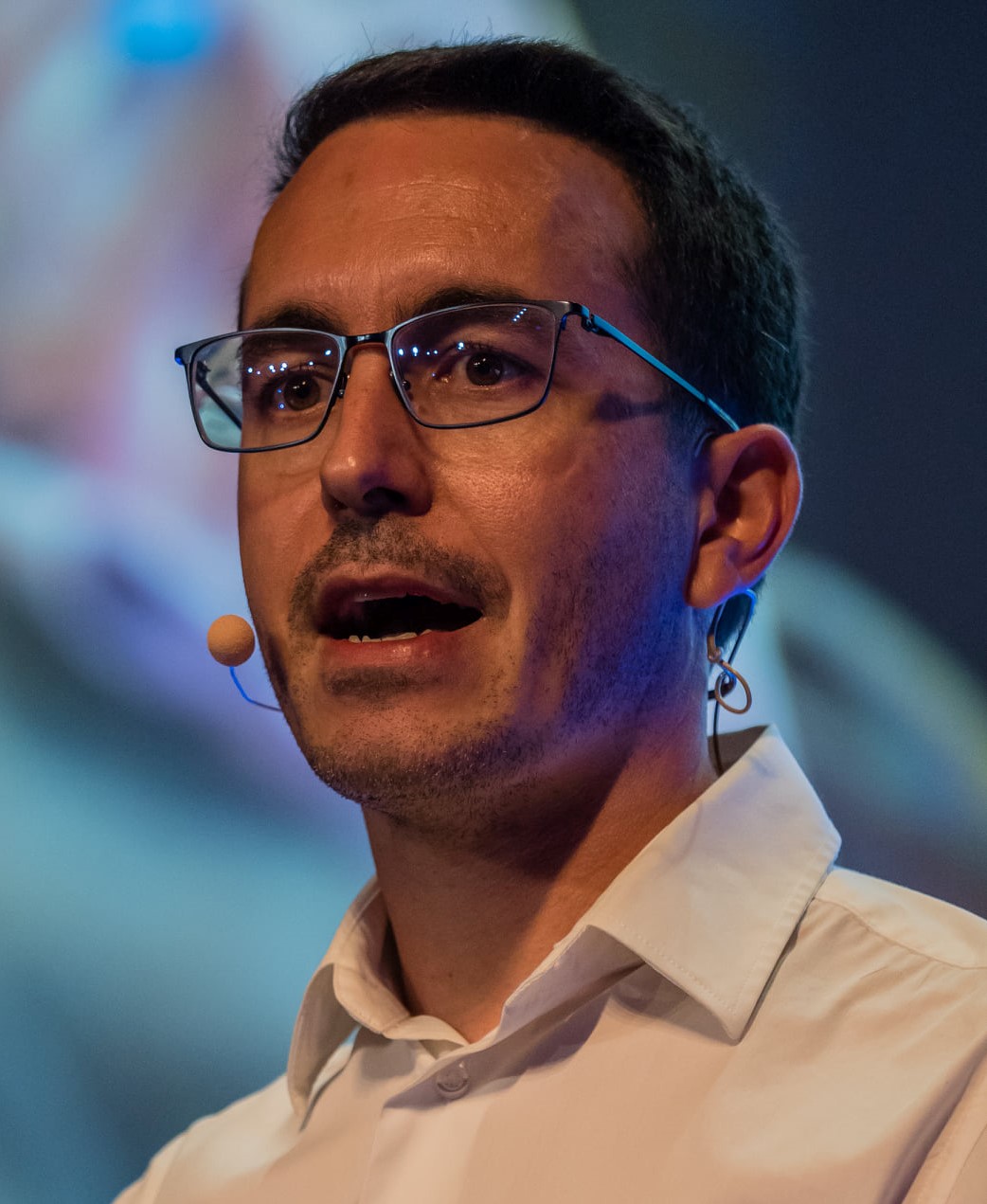}}]{Aythami Morales Moreno}
received his M.Sc. degree in Electrical Engineering in 2006 from Universidad de Las Palmas de Gran Canaria. He received his Ph.D. degree in Artificial Intelligence from La Universidad de Las Palmas de Gran Canaria in 2011. He performs his research works in the BiDA Lab – Biometric and Data Pattern Analytics Laboratory at Universidad Autónoma de Madrid, where he is currently an Associate Professor (CAM Lecturer Excellence Program). He is a member of the ELLIS Society (European Laboratory for Learning and Intelligent Systems).
\end{IEEEbiography}
\vskip -2\baselineskip plus -1fil
\begin{IEEEbiography}[{\includegraphics[width=1.1in,height=1.5in,clip,keepaspectratio]{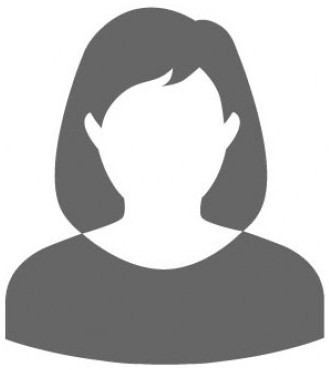}}]{Els J. Kindt} obtained her master degree in law at KU Leuven, Belgium and an LL.M in law at UGA, USA and is affiliated  as a post-doc legal researcher and associate professor with respectively the  Centre for  IT and  IP Law (CITIP) of KU Leuven, Belgium and eLaw of Universiteit Leiden, the Netherlands.  Besides participating in several  national and European research projects (BioSec, Turbine, Fidelity, Eksistenz, Beat, iMars, etc.) as principal investigator, she has also been teaching, in particular European Privacy and Data Protection law and Electronic contracts law at KU Leuven, Brussels, Biometrics and the law at Unidistance, Switzerland and is member of various editorial boards, including EDPL.
She is recognized as a specialist on legal aspects of biometric technologies (see also Privacy and Data Protection Issues of Biometric Applications. A Comparative Legal Analysis,at Springer) and set up the Biometric Law Lab at CiTiP, consisting of currently 7 legal researchers in biometrics. 
\end{IEEEbiography}






\end{document}